\documentclass[conference, 10pt, letterpaper]{IEEEtran}
\IEEEoverridecommandlockouts

\usepackage{cite}
\usepackage{amsmath,amssymb,amsfonts}
\usepackage{algorithmic}
\usepackage{graphicx}
\usepackage{textcomp}
\usepackage{xcolor}
\usepackage{tikz}
\usepackage[ruled,vlined]{algorithm2e}
\def\BibTeX{{\rm B\kern-.05em{\sc i\kern-.025em b}\kern-.08em
    T\kern-.1667em\lower.7ex\hbox{E}\kern-.125emX}}

\newcommand\copyrighttext{%
  \footnotesize 978-1-5386-5541-2/18/\$31.00~\copyright2020 IEEE. Personal use of this material is permitted. Permission from IEEE must be obtained for all other uses, in any current or future media, including reprinting/republishing this material for advertising or promotional purposes, creating new collective works, for resale or redistribution to servers or lists, or reuse of any copyrighted component of this work in other works. DOI: }
\newcommand\copyrightnotice{%
\begin{tikzpicture}[remember picture,overlay]
\node[anchor=south,yshift=10pt] at (current page.south) {\fbox{\parbox{\dimexpr\textwidth-\fboxsep-\fboxrule\relax}{\copyrighttext}}};
\end{tikzpicture}%
}

\begin{document}

\title{Exploring time-series motifs through DTW-SOM}

\author{\IEEEauthorblockN{Maria Inês Silva}
\IEEEauthorblockA{\textit{Nova Information Management School (NOVA IMS),} \\
\textit{Campus de Campolide, Universidade Nova de Lisboa,} \\ 
1070-312 Lisboa, Portugal \\
\texttt{d20170088@novaims.unl.pt}}
\and
\IEEEauthorblockN{Roberto Henriques}
\IEEEauthorblockA{\textit{Nova Information Management School (NOVA IMS),} \\
\textit{Campus de Campolide, Universidade Nova de Lisboa,} \\ 
1070-312 Lisboa, Portugal \\
\texttt{roberto@novaims.unl.pt}}
}

\maketitle

\copyrightnotice

\begin{abstract}
Motif discovery is a fundamental step in data mining tasks for time-series data such as clustering, classification and anomaly detection. Even though many papers have addressed the problem of how to find motifs in time-series by proposing new motif discovery algorithms, not much work has been done on the exploration of the motifs extracted by these algorithms. In this paper, we argue that visually exploring time-series motifs computed by motif discovery algorithms can be useful to understand and debug results.

To explore the output of motif discovery algorithms, we propose the use of an adapted Self-Organizing Map, the DTW-SOM, on the list of motif's centers. In short, DTW-SOM is a vanilla Self-Organizing Map with three main differences, namely (1) the use the Dynamic Time Warping distance instead of the Euclidean distance, (2) the adoption of two new network initialization routines (a random sample initialization and an anchor initialization) and (3) the adjustment of the \textit{Adaptation} phase of the training to work with variable-length time-series sequences.

We test DTW-SOM in a synthetic motif dataset and two real time-series datasets from the UCR Time Series Classification Archive \cite{UCRArchive2018}. After an exploration of results, we conclude that DTW-SOM is capable of extracting relevant information from a set of motifs and display it in a visualization that is space-efficient.
\end{abstract}


\section{Introduction}

\begin{figure}[htbp]
\centerline{\includegraphics[width=0.35\textwidth]{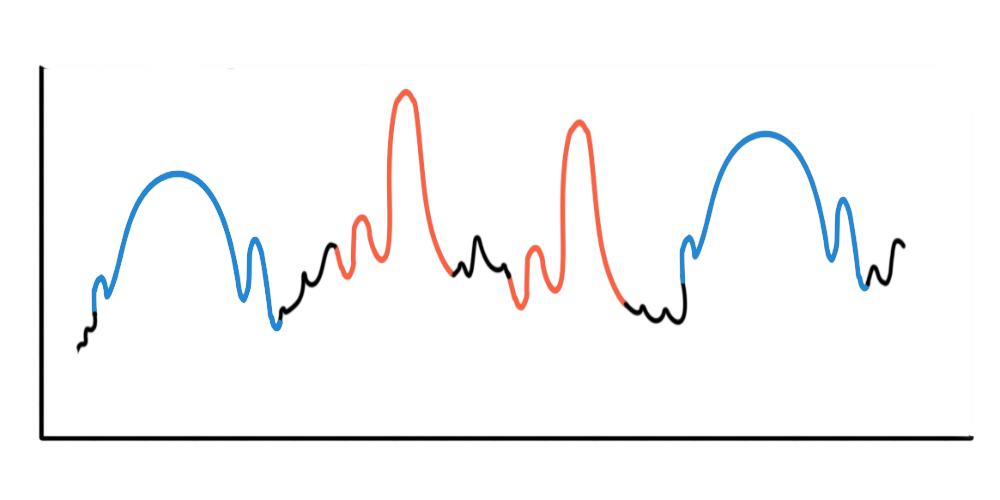}}
\caption{Toy example of two different motifs, each with two highly conserved subsequences.}
\label{fig-motif}
\end{figure}

In the last decade, motif discovery has become a fundamental step in many data mining tasks for time-series data, such as clustering, classification or anomaly detection. In general, a time-series motif corresponds to a over-represented segment of a time-series and thus motif discovery involves extracting all (or a specific subset) of these over-represented segments \cite{mueen_time_2014}. Figure \ref{fig-motif} illustrates an example of two motifs built from dummy data.

Due to its relevance, many methods and strategies have been proposed to tackle motif discovery. However, the step of exploring and visualizing motifs, which can be useful to understand results of downward tasks, has not received as much attention. To the best of our knowledge, papers that address this question focus only on visualizing the actual time-series subsequences that belong to each individual motif \cite{hao_visual_2012, balasubramanian_flexible_2013, senin_grammarviz_2014}. We argue that, even though exploring individual motifs can help to understand the individual patterns, these methods cannot provide information about the overall relationships between the extracted motifs. In other words, they are not ideal to answer questions such as: Are motifs similar to each other? Can we define clusters of motifs? Additionally, exploring individual clusters is not tractable in the cases where a high number of motifs is extracted. 

In this paper, we propose the use of a widely-studied method for feature reduction and visualization, the Self-Organizing Map (SOM) \cite{Kohonen2001}, to explore the centers of motifs extracted by any desired motif discovery algorithm. Taking into account that these centers are time-series subsequences, with possibly variable lengths and multiple dimensions, we adapted the original SOM algorithm to apply the \textit{Dynamic Time Warping} (DTW) distance \cite{berndt_using_1994} as its similarity metric and added two specific initialization routines for the SOM network.

The rest of the paper is organized as follows: section 2 introduces academic work related to (1) motif discovery and (2) Self-Organizing Maps, section 3 describes our own implementation of the \textit{Dynamic Time Warping Self-Organizing Map} (DTW-SOM), section 4 presents the experimental setup and reports the results obtained on three different datasets and, finally, section 5 concludes this paper and discusses future work.

\section{Related work}

In this section, we'll cover two areas which, although seemingly unrelated, serve as basis for this paper - motif discovery and self-organizing maps.

\subsection{Motif Discovery}

Despite, in general, the concept of motifs being associated with significant time-series segments, there are two main definitions of motifs that vary on the way they set the concept of "significance"  \cite{mueen_time_2014}. Similarity-based motifs focus on the similarity of the time-series segments and thus this definition results in highly similar motifs. On the other hand,  support-based motifs focus on the repetition of the segments throughout the time-series and thus this definition leads to highly frequent motifs.

In addition to the concept of significance, there are additional constraints a group of time-series segments must meet to be considered a motif \cite{tanaka_discovery_2005}. The first is a behavior constraint, which determines that segments in a motif should have the same general behavior, even if some level of noise or time and amplitude shifts are allowed \cite{torkamani_survey_2017}.  The second, the non-overlapping constraint, aims to avoid trivial matchings \cite{lin_finding_2002} by setting that motif segments cannot overlap in time. The third is a distance constraint, which restricts all motif's segments to have a distance smaller than a radius $R$ to the center of the motif (i.e., the segment that represents that motif). Note that in most motif discovery methods, the radius is a parameter that needs to be set by the user. Finally, when the task is to extract a set of motifs instead of the most significant motif, there is an additional constraint with the goal of avoiding the extraction of motifs with overlapping members. This last constraint states that all the motifs' centers must have a distance higher than 2 radius, $2R$.

Independently of the exact definition, motif discovery is a computationally intensive task as it involves computing distances between all possible pairs of time-series subsequences. Therefore, much of the work related to motif discovery algorithms has aimed at making the search more efficient. One of the most common techniques to reduce the search space is to convert the original time-series into a lower-dimensional representation where the distance between subsequences in the original representation is approximately maintained \cite{mueen_time_2014}.  With this strategy, motif candidates can be extracted in the low-dimensional representation (which is more efficient) and the final set of motifs can be computed with the real distance on the smaller set of candidates. In motif discovery, the \textit{Symbolic Aggregate approXimation} (SAX) representation \cite{lin_experiencing_2007} is the most used. This representation first extracts sliding windows from the original time-series and then converts each sliding window into a fixed-sized sequence of characters.  Thus, the time-series is split into a list of "words".

An important part of any motif discovery is the choice of distance. Most authors use the Euclidean distance \cite{das_rule_1998} or the \textit{Dynamic Time Warping} (DTW) distance \cite{berndt_using_1994} for comparing subsequences. The advantage of the Euclidean is its efficiency, which allows a faster comparison of motif candidates. However, this efficiency comes with a loss of flexibility as it is not robust to time-shifts, distortions, differences in phase and variable-length sequences. On the other hand, the DTW distance finds the optimal time alignment between the subsequences that are being compared. Thus, it is much less efficient but it can adapt to the shifts discussed before.

In terms of the algorithms for motif discovery, there are two main types, namely fixed-length and variable-length. In the fixed-length algorithms, users need to provide the length of window and that parameter is used to extract all the possible subsequences. Thus, all motifs contain subsequences of the same size. The MK exact algorithm \cite{mueen_exact_2009}, the motif extraction from the Matrix Profile \cite{yeh_matrix_2016} and the EMMA-SAX algorithm \cite{lin_finding_2002} are all examples of methods that extract fixed-length motifs.

The variable-length algorithms are a bit more flexible and don't required the user to set the window size beforehand. However, this is a much harder problem as the search space is bigger and more complex. Most variable-length algorithms solve this problem by applying a fixed-length algorithm to a range of window sizes and then choosing the most representative motifs based on their motif definition and motif ranking schemes \cite{torkamani_survey_2017}. The work of \cite{nunthanid_parameter-free_2012} and \cite{gao_exploring_2018} are two examples of this approach. On the other hand, Lin's grammar-based approach \cite{lin_finding_2010} and Tanaka's EMD algorithm \cite{tanaka_discovery_2005} already take into consideration variable-length motif during the search process by adapting the way subsequences are represented in the low-dimensional representation.

\subsection{Self-organizing map}

Self-organizing Maps (SOM) were proposed by Tuevo Kohonen at the beginning of the 1980s \cite{Kohonen2001}, and constitute the product of his work on associative memory and vector quantization. The SOM's basic idea is to map high-dimensional data onto a low-dimensional discrete feature map, maintaining the relations between data patterns \cite{Henriques2010}. Its main objective is to "extract and illustrate" the essential structures from a dataset through a map resulting from an unsupervised learning process \cite{Kaski1996,Kaski1998} and thus it can be used at the same time for visualization and exploration of data and for clustering.

SOM is also considered a good method for extracting data patterns and associations when the extraction of information becomes a challenging task due to the number of parameters or the use of a multidimensional dataset.

Usually, SOM maps the original high-dimensional data to a discrete feature map with one, two, or three-dimensions, although 2-dimensions are the most common. The grid formed by the units or neurons is usually referred to as output space, as opposed to input space, which is the original space \cite{Henriques2009}. When the output space is 2-dimensional, it is usually formed by a rectangular or hexagonal grid of units \cite{Kohonen2001}. 

Each unit of the SOM, is represented by a vector \(m_i=[m_{i1}...,m_{in}]\) of dimension \(n\), where \(n\) equals the dimension of the input space. In the training phase, a given training pattern \(x\) is presented to the network, and the closest unit is selected. This unit is called the best-matching unit (BMU). The unit’s vector values and those of its neighbors are then modified in order to get closer to the data pattern $x$:

\begin{equation}\label{eu_som}
    m_i=m_i+\alpha(t) h_{ci}(t)\|x-m_i\|
\end{equation}

where $\alpha(t)$  is the learning rate at time $t$, and $h_{ci}(t)$ is the neighborhood function centered in unit $c$, and $i$ identifies each unit. To allow SOM to converge to a stable solution, both the learning rate $\alpha(t)$ and the neighborhood radius $h_{ci}(t)$ should decrease to zero during training. Usually these parameters decrease in a linear fashion but other functions can be used. Additionally, the update of both parameters can be done after each individual data pattern is presented to the network (iteration) or after all the data patterns have been presented (epoch). The former case is known as sequential training and the latter is usually known as batch training. The sequential algorithm pseudo-code is presented bellow, in algorithm \ref{som_training}.

\begin{algorithm}
\SetAlgoLined
\KwIn{$X = \{ x_1,x_2,…x_n \}$: training patterns\newline
      $W = \left( w_{ij} \right) \in \mathbb{R}^{p \times q}$: SOM network's units\newline
      $\alpha \in ]0,1[$: initial learning rate\newline
      $r \in \mathbb{R}$: initial neighborhood radius}
    Let $h (w_{ij},w_{mn},r)$ be the neighborhood function\newline
    \Repeat{$\alpha = 0$}{
        \For{$k=1$ to $n$)}{
            \ForAll{$w_{ij} \in W$}{$d_{ij}=\|x_k- x_{ij}\|$}
            Select the unit that minimizes $d_{ij}$ as the winner $w_{win}$
            
            Update $w_{ij}\in W: \; w_{ij}=w_{ij} + \alpha h(w_{win},w_{ij},r)\|x_k-w_{ij}\|$
        }
        Decrease the value of $\alpha$ and $r$
    }
\label{som_training}
\caption{SOM Sequential Training}
\end{algorithm}

For each randomly selected training pattern presented to the network, the BMU (i.e. its closest unit) is found. The BMU is then updated according to the weights of the training pattern and the learning rate. Initially this learning rate is high allowing bigger adjustments of the units. The unit’s mobility will decrease proportionality with the decrease of the learning rate. Based on the neighborhood rate, a group of surrounding units is also moved closer to the training pattern. There are several ways to visualize the SOM and improve the understanding of the data patterns \cite{Vesanto1999}. Two of the most important visualization tools are the component planes \cite{Kohonen2001} and the U-matrix \cite{Ultsch1990}.

In a component plane, each unit is colored according to the weight of each variable in the SOM. Through the analysis of the component planes, data distribution can be evaluated.  For  instance,  it  is  quite  simple  to  identify  variables  which  are correlated (their component planes will have the same shape), and it is also possible to have  an  improved  understanding  of  the  contributions  of  each  variable  to  the  SOM. By comparing  two  or  more  component  planes,  one  can  visually  identify  correlations between variables, both globally and at a local scale.

The  U-matrix  is  one  of  the  most  used  methods  to  visualize  SOM \cite{Ultsch1990}. U-matrices  are  computed  by  finding  the  distances  in  the  input  space  of neighboring units in the output space. There are two ways to visualize a U-matrix. The most common is to use a color code to depict distances, corresponding to the values of the U-matrix.  Usually a grey-scale  is  used,  with the highest value being represented with black and the lowest with  white.  Another possibility is to plot these distances in the form of a 3D landscape with mountains and valleys. A mountain region indicates  large  distances  between  units,  while  low  distances  between  the  units  form valleys.

Some of the challenges in applying SOM to motif discovery in time-series are related to the fact that not all dimensions are equally relevant and the different size of motifs. Some work has been developed to overcame these problems \cite{wang2013,bassani2015dimension, Brito_2018}. One example is the Local Adaptive  Receptive Field Dimension Selective Self-organizing Map (LARFDSSOM), proposed  in  \cite{bassani2015dimension}, where the application of different weights for each  input dimension and for each cluster is proposed, as well as the use of a Time-Varying Structure of the SOM. In \cite{Brito_2018} the authors propose VILMAP to allow the use of different sizes of samples and consequently the discovery of Motifs with different lengths.

SOM can, as shown in the previous references, be used to identify and extract time-series motifs. In this paper, we  propose the use of this technique not to perform such tasks but to analyze the motifs extracted by others motif discovery algorithms.  

\section{DTW-SOM}

As explained in previous sections, the goal of the DTW-SOM method is to serve as an auxiliary tool in motif discovery and, as such, it needs to be flexible enough to deal with a different range of motif discovery methods and definitions. Particularly, it should:

\begin{enumerate}
    \item Be able to compare time-series segments.
    \item Receive all types of motifs, meaning fixed-length and variable-length motifs.
    \item Represent multi-dimensional motifs (i.e., motifs extracted from a multi-dimensional time-series).
    \item Take into account an given order of significance in the motifs of a provided set of "most significant" motifs.
    \item Provide a visualization from which a user can get insights into the general relationships between the motifs and their shapes.
\end{enumerate}

To achieve these goals, we extended the vanilla batch-training SOM algorithm that was implemented in the PyClustering python package \cite{Novikov2019} to process variable-length multidimensional time-series subsequences. The idea is that when someone wishes to investigate a set of motifs previously extracted, he just needs to collect the center subsequences of each motif and feed that list of centers to the DTW-SOM. For readability, from now on, we'll refer to these center subsequences that are the input to the DTW-SOM as \textit{patterns}.

Firstly, we added two network initialization routines, namely a random sample initialization and an anchor initialization. In the random sample initialization, each unit is randomly assigned to one input pattern in such a way that no two units are exactly the same. In other words, a random sample is taken from the input patterns to initialize the network. In the anchor initialization, a list of patterns must be provided by the users (the anchors) and each anchor will be set to a single unit. Note that the provided list cannot have more anchors than units and, in the case of an input with less anchors then units, a random sample will be taken from all the input patterns to initialize the remaining units. In the case of having less anchors than units, we tried to spread out the first anchors (as we are assuming that they are ordered by significance) by first filling the diagonals of the network and just after that filling the rest of the network.

Note that the anchor initialization was designed to address the fourth requirement presented above. If the motif discovery method provides an ordering of the motifs or a subset of most significant motifs, one can use these motifs as the anchors and thus the DTW-SOM will more easily focus on these more important motifs.

The second adjustment we implemented to the vanilla SOM was the swap of the distance function from the Euclidean distance to the DTW distance, which allows us to process patterns with different lengths and with multiple dimensions. Finally, the last big change implemented was in the \textit{Adaptation} phase of the training. Since we are comparing patterns and units with variable lengths, we had to adapt the way each unit's values are updated. When computing the distance between two segments, the DTW distance finds an optimal alignment in time between the segments and uses this optimal match to compute the distance between individual points along the segments. Thus during the adaptation phase of training we leverage this matching to guide the update of the BMU's vector values.

\begin{figure}[htbp]
\centerline{\includegraphics[width=0.35\textwidth]{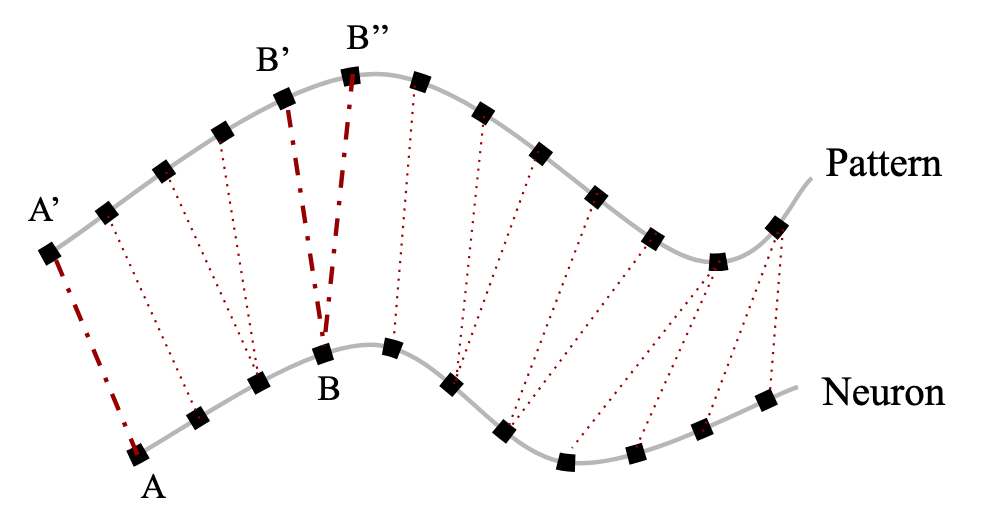}}
\caption{Example of the time-alignment between an input pattern and an unit computed by the DTW algorithm. The black points are the actual time-series observations while the red and dashed lines represent the alignment matching returned by the DTW algorithm.}
\label{fig-dtw}
\end{figure}

As an example, figure \ref{fig-dtw} shows the DTW alignment between an input pattern and its BMU. In the DTW-SOM, the vector values of an unit are simply the sequence of time-series observations as presented in figure. During \textit{Adaptation}, every vector value needs to move slightly closer to the pattern as defined by equation \ref{eu_som}. In this case, the first BMU's value, which is the point marked as $A$, will move closer to the its matching point in the pattern, which is the point $A'$.

Due to warping, we can have cases where one point in the BMU can be matched to more than one point in the pattern, as is the case with points $B$, $B'$ and $B''$. In these cases, equation \ref{eu_som} needs to be changed to equation \ref{dtw_som_eq}:

\begin{equation}\label{dtw_som_eq}
    w_i=w_i+\alpha(t) h_{cw}(t)\left( \frac{1}{n}\sum_{j \leq n} x_j - w_i \right)
\end{equation}

where $i$ is the index of the unit's sequence, $t$ is the training epoch, $\alpha(t)$ is the learning rate at epoch $t$, $h_{cw}(t)$ is the neighborhood function, $n$ is the number of pattern points that where matched to the BMU's vector value $w_i$ and the $x_j$ are the pattern point's values.

\section{Evaluation}

To test DTW-SOM, we did three experiments. In the first, we generated synthetic motif data and explored visually the results. In the second, we used a widely used classification dataset from the UCR Time Series Classification Archive \cite{UCRArchive2018}, the \textit{GunPoint} dataset \cite{Ratanamahatana2004}. Particularly, we adapted the dataset to be suitable for the motif discovery task, we used the Matrix Profile \cite{yeh_matrix_2016} to efficiently extract all the motifs and we explored the resulting motifs with our method. Finally, in the third experiments, we used the same approach as in the second experiments, but used a dataset from the UCR Time Series Classification Archive with more classes and more extracted motifs, namely the \textit{UWaveGesture} dataset \cite{liu_uwave_2009}.

\subsection{Experiment with the synthetic motif dataset}
 
To build this synthetic dataset, the idea was to create a dataset of motifs centers that formed 3 clear clusters. If we were able to detect these clusters in the final visualization, then the DTW-SOM was working as expected. The synthetic dataset included 180 motif centers which were generated using the following heuristic:

\begin{enumerate}
    \item We chose three general behaviors sequences for the clusters, namely, low-middle-high, high-middle-low and middle-middle-middle.
    \item For each behavior (low, middle and high), we defined intervals from which we could sample points exhibiting that behavior. Particularly, the low interval was $[-3, -1.5]$, the middle interval was $[-0.5, 0.5]$ and the high interval was $[1.5, 3]$.
    \item For each motif center, we set the length of which behavior by randomly selecting an integer between 5 and 10. In other words, for each motif center, we'll sample three integers that define the lengths of each of its behavior subsequences and the sum of those integers will be the total length of that motif center.
    \item For each motif center, we create its time-series sequence by sampling values from the predefined behavior intervals. As an example, if we were creating a motif center for the low-middle-high cluster and if we had previously sampled the behavior lengths $(3, 7, 4)$, then we would sample three values from the interval $[-3, -1.5]$, seven values from the interval $[-0.5, 0.5]$ and four values from the interval $[1.5, 3]$.
\end{enumerate}

\begin{figure}[htbp]
\centerline{\includegraphics[width=0.35\textwidth]{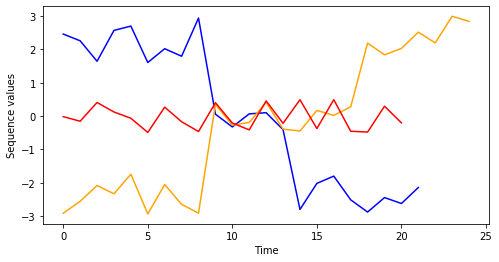}}
\caption{Plot with three examples of generated motif centers, one for each cluster. The orange belongs to the low-middle-high cluster, the blue to the high-middle-low cluster and the red to the middle-middle-middle cluster}
\label{fig-syn-motifs}
\end{figure}

Figure \ref{fig-syn-motifs} shows some examples of the generated sequences. After generating the dataset, we build a DTW-SOM network with a $3X3$ layout and default parameters and we trained it during 30 epochs. We also tested the random sample initialization and the anchor initialization (using one motif from each cluster as the anchors). We noted that even though the random sample initialization was able to obtain the desired clusters, the results among different training runs were much more unstable. On the other hand, the anchor initialization converged to the desired clusters much more consistently and different runs did not change too much the results.

\begin{figure}[htbp]
\centerline{
\includegraphics[width=0.21\textwidth]{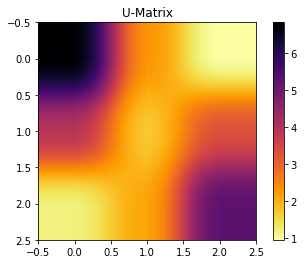}
\includegraphics[width=0.185\textwidth]{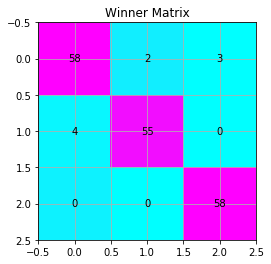}
}
\caption{U-matrix and Winner Matrix obtained from the DTW-SOM trained on the synthetic motifs and using an anchor initialization.}
\label{fig-syn-umatrix}
\end{figure}

\begin{figure}[htbp]
\centerline{\includegraphics[width=0.45\textwidth]{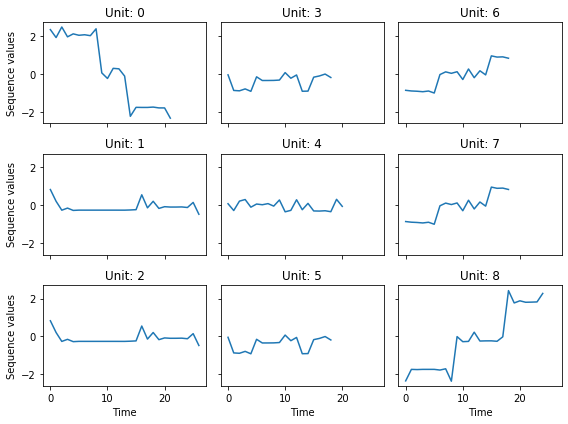}}
\caption{Sequence values of the units after training the DTW-SOM with synthetic motifs. The position of each unit's plot in the grid is consistent with its position in the network. Thus, the U-Matrix and the Winner matrix plots are consistent with this grid plot.}
\label{fig-syn-units}
\end{figure}

Figure \ref{fig-syn-umatrix} shows the U-Matrix and the Winner Matrix obtained with the anchor initialization. Note that the winner matrix only encodes the number of input patterns that had each units as its BMU. We also plotted the sequence values of the nine units, which can be observed in figure \ref{fig-syn-units}. From both plots, we can see that the diagonal of the network captures almost all the motifs as expected and that the rest of the units capture some other patterns between the middle-middle-middle cluster and the low-middle-high cluster.

\subsection{Experiment with the \textit{GunPoint} dataset}

 The \textit{GunPoint} dataset \cite{Ratanamahatana2004} was built from the hand motion of two actors that are performing two different actions - the first is to draw a gun (which corresponds to the class "Gun") and the second is to point a finger (which corresponds to the class "Point"). The time-series correspond to the measurements in the x-axis of tracking the centroid of the actor's right hand. Because the setup was a classification task, we have a train and a test sets with 50 and 150 time-series sequences, respectively. Each time-series sequence includes the whole action of either the gun draw or the finger pointing and every sequence has a length of 150.
 
\begin{figure}[htbp]
\centerline{
\includegraphics[width=0.2\textwidth]{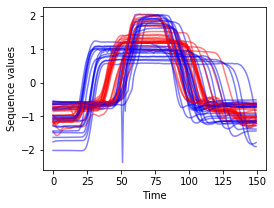}
\includegraphics[width=0.28\textwidth]{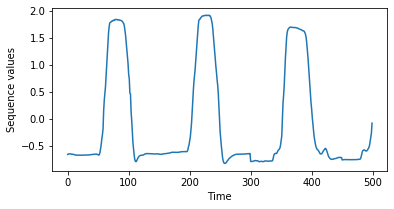}
}
\caption{\textbf{Right plot:} original sequences from \textit{GunPoint}'s train set. The colors indicate the different labels, "Gun" and "Point". \textbf{Left plot:} first 500 observations of the time-series built as a concatenation of the fixed-size sequences from \textit{GunPoint}'s train set.}
\label{fig-gp-time-series}
\end{figure}

 Due to the nature of this dataset, we had to adapt it to the task of motif discovery. For simplicity, we concatenated the 50 sequences from the train set into a single time-series and use this time-series as input to the motif discovery algorithm. Figure \ref{fig-gp-time-series} includes a visualization with the original sequences and a subset of the final concatenated time-series.
 
The algorithm we used was the one based on the Matrix Profile \cite{yeh_matrix_2016}. Firstly, since we had the original lengths of the sequences, we could pose the problem as a fixed-length motifs discovery. Secondly, the Matrix Profile is known to be very efficient and, thirdly, its parameters are few and easy to tune.

Note that this method expects to receive as input the max number of motifs to find and thus this is actually a k-motif algorithm. However, if we set this parameter larger than expected (e.g. 1000 in our case), then the algorithm will return all the motifs it can find. This is exactly what we did and the algorithm managed to extract 25 motifs.

\begin{figure}[htbp]
\centerline{
\includegraphics[width=0.21\textwidth]{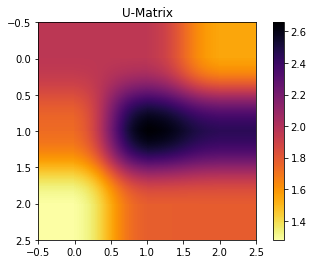}
\includegraphics[width=0.185\textwidth]{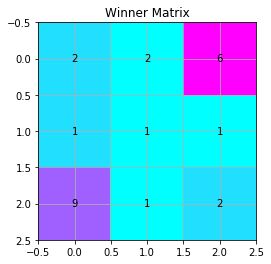}
}
\caption{U-matrix and Winner Matrix obtained from the DTW-SOM trained on the motifs computed from the concatenated time-series of \textit{GunPoint} sequences.}
\label{fig-gp-umatrix}
\end{figure}

\begin{figure}[htbp]
\centerline{\includegraphics[width=0.45\textwidth]{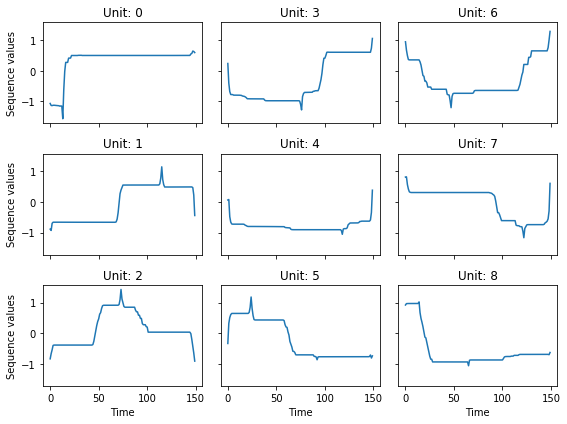}}
\caption{Sequence values of the units after training the DTW-SOM with motifs from the \textit{GunPoint} dataset. The position of each unit's plot in the grid is consistent with its position in the network. Thus, the U-Matrix and the Winner matrix plots are consistent with this grid plot.}
\label{fig-gp-units}
\end{figure}

Finally, we built a DTW-SOM network with a $3X3$ layout, default parameters and a random sample initialization. We then rained that network with the list of the motifs' centers during 50 epochs. Figure \ref{fig-gp-umatrix} shows the U-Matrix and the Winner Matrix obtained from this DTW-SOM network and Figure \ref{fig-gp-units} has the plots of the units' sequence values.

In this dataset, we can observe that DTW-SOM was capable of extracting some interesting information about the original 25 motifs computed with the Matrix Profile algorithm. Firstly, we can see two clear clusters around the units 2 and 6 (i.e. in the down-left and the up-right corners of the network).

Unit 2 corresponds to a pattern of raising the hand (either with a gun or not) and lowering the hand. Its neighbors, units 1 and 5, have the same pattern and since they were the BMU of a single input motif, they are essentially a cluster with unit 2.

Unit 6, on the other hand, has the pattern of lowering the hand and raising it again. It corresponds to the end of one of the original sequences and the start of the next one. Units 3 and 7 also have the lowering-raising patterns, however unit 3 has more time of the raising while 7 has more time of lowering.

Finally, units 0, 4 and 8 have their own specific pattern. Unit zero seems to have the original sequences of raising the hand a bit lower than the rest of sequences. In other words, they are the flat sequences in figure \ref{fig-gp-time-series} that peak at the value 1. Unit 4 is the BMU of a single motif and encodes the "no action" pattern. Unit 8 has a quicker lowering pattern and thus encodes the end of the sequences with a quicker movement.

\subsection{Experiment with the \textit{UWaveGesture} dataset}

\begin{figure}[htbp]
\centerline{
\includegraphics[width=0.2\textwidth]{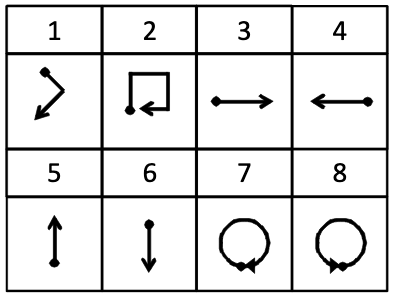}
\includegraphics[width=0.2\textwidth]{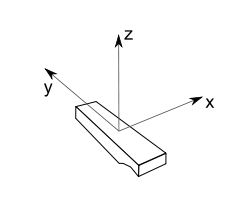}
}
\caption{\textbf{Right plot:} Gesture vocabulary and related labels used in the \textit{UWaveGesture} dataset, as presented in \cite{liu_uwave_2009}. \textbf{Left plot:} Positioning of the accelerometer axis in the Wii remote.}
\label{fig-uwg-labels}
\end{figure}

The \textit{UWaveGesture} dataset \cite{liu_uwave_2009} corresponds to accelerometer recordings of right-hand gestures performed with the Wii remote. The dataset was built from eight participants doing eight specific gestures, which are presented on the left side of Figure \ref{fig-uwg-labels}. The remote collects acceleration measurements from its three axis, as presented in the right side of Figure \ref{fig-uwg-labels}, and the UCR Time Series Classification Archive has one time-series dataset for each axis.

\begin{figure}[htbp]
\centerline{
\includegraphics[width=0.45\textwidth]{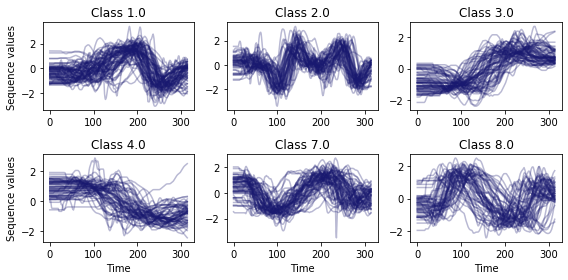}
}
\centerline{
\includegraphics[width=0.45\textwidth]{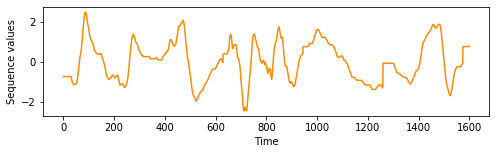}
}
\caption{\textbf{Blue plots:} Original sequences sampled from \textit{UWaveGesture}'s train set, split by the gesture class. \textbf{Orange plot:} First 1600 observations of the time-series built as a concatenation of the fixed-size sequences from \textit{UWaveGesture}'s train set}
\label{fig-uwg-time-series}
\end{figure}

For this experiment, we chose to use the dataset with the x-axis recordings, which corresponds to the lateral movements of the Wii remote. Because in this axis the gestures 5 and 6 have a zero acceleration (and thus are only noise), we excluded the subsequences with these classes. In order to accelerate computation, we also sampled 400 sequences from the train set. Finally, similarly to the previous experiment, we concatenated the sampled sequences into a single time-series from which we could extract motifs. Figure \ref{fig-uwg-time-series} contains the original sequences, split by the gesture, and a subset of the time-series that resulted from the sequences' concatenation.

\begin{figure}[htbp]
\centerline{
\includegraphics[width=0.21\textwidth]{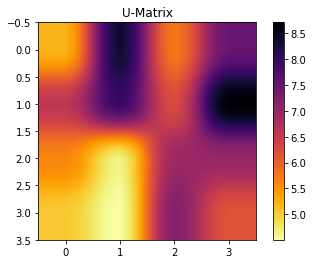}
\includegraphics[width=0.185\textwidth]{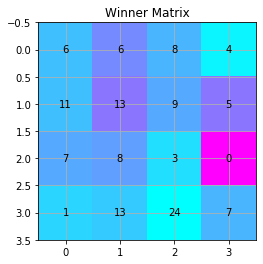}
}
\caption{U-matrix and Winner Matrix obtained from the DTW-SOM trained on the motifs computed from the concatenated time-series of \textit{UWaveGesture} sequences.}
\label{fig-uwg-umatrix}
\end{figure}

\begin{figure}[htbp]
\centerline{\includegraphics[width=0.45\textwidth]{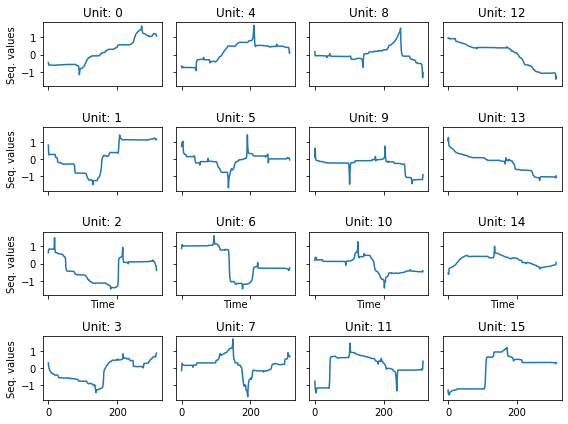}}
\caption{Sequence values of the units after training the DTW-SOM with motifs from the \textit{UWaveGesture} dataset. The position of each unit's plot in the grid is consistent with its position in the network. Thus, the U-Matrix and the Winner matrix plots are consistent with this grid plot.}
\label{fig-uwg-units}
\end{figure}

In this experiment, we used again the Matrix Profile \cite{yeh_matrix_2016} with a larger than expected max number of motifs in order to extract all the fixed-length motifs. In this dataset, the algorithm extracted 125 motifs. We then trained a DTW-SOM network with a $4X4$ layout, using the default parameters, a DTW maximum window of 100 (to limit the warping level) and a random sample initialization, which resulted in the U-Matrix, the Winner Matrix and the units shown in Figure \ref{fig-uwg-umatrix} and \ref{fig-uwg-units}.

From the U-matrix we can distinguish different regions of the DTW-SOM network. Unit 0, which forms its own cluster, has a simple shape similar to the third gesture. This is a simple gesture of moving the Wii remote to the right. Units 12 and 13 are far from their neighboring units and have a shape similar to the fourth gesture, or the gesture of moving the Wii remote to the left. These are the simplest gestures involving lateral movement of the Wii remote and it is expected that our motif detection algorithm would pick on these shapes.

The visually biggest cluster in the U-matrix is centered at units 6 and 7. These units have a shape similar to the first and eighth gestures. These gestures are made of a right lateral movement followed by a left lateral movement. And so we can see that the motif detection algorithm is being capable of extracting more complex shapes.

Even though they don't form a visually striking cluster in the U-matrix, units 1, 2 and 5 have very similar shapes and are the only units with a shape consistent with the left-right lateral movement present in the seventh gesture.

Interestingly, unit 15 forms its own cluster in the lower-right corner of the DTW-SOM network, but it has a shape consistent with the third gesture. In other words, we have two clusters in the network, one in unit 0 and another in unit 15, with the same shape. This is due to the random initialization. Because these two similar motifs were randomly assigned to far away places in the network, they had no option but to form two independent clusters.

\section{Conclusion}

In this paper, we argue that visually exploring the time-series motifs computed by motif discovery algorithms can be useful to understand and debug results. To the best of our knowledge, no other papers investigate the problem of exploring relationships between motifs and answering questions such as: Are motifs similar to each other? Can  we  define  clusters  of  motifs?

To conduct these investigations, we propose the use of an adapted Self-Organizing Map on the list of motif's centers. We called the adapted method DTW-SOM and the main changes are (1) the use the Dynamic Time Warping distance to compute distances between the units and the input patterns, (2) the introduction of two new network initialization routines and (3) the adjustment of the \textit{Adaptation} phase of the training to work with variable-length time-series sequences.

We tested DTW-SOM in a synthetic motif dataset and two real time-series datasets called \textit{GunPoint} and \textit{UWaveGesture}, respectively. From an exploration of results, we can conclude that DTW-SOM is capable of extracting relevant information from a set of motifs and display it in a space-efficient way. During the experiment with the synthetic dataset, we observed that the random sample initialization was not as robust as the anchors initialization. Additionally, this random initialization can also lead to the creation of distinct clusters that have the same shapes, which is not optimal.  Thus, as future work, we propose an investigation on more robust initialization schemes to cover the case when the user does not wish to provide anchors.

\bibliographystyle{./bibliography/IEEEtran}
\bibliography{./bibliography/IEEEabrv,./bibliography/references,./bibliography/robertoRef}

\begin{thebibliography}{10}
\providecommand{\url}[1]{#1}
\csname url@samestyle\endcsname
\providecommand{\newblock}{\relax}
\providecommand{\bibinfo}[2]{#2}
\providecommand{\BIBentrySTDinterwordspacing}{\spaceskip=0pt\relax}
\providecommand{\BIBentryALTinterwordstretchfactor}{4}
\providecommand{\BIBentryALTinterwordspacing}{\spaceskip=\fontdimen2\font plus
\BIBentryALTinterwordstretchfactor\fontdimen3\font minus
  \fontdimen4\font\relax}
\providecommand{\BIBforeignlanguage}[2]{{%
\expandafter\ifx\csname l@#1\endcsname\relax
\typeout{** WARNING: IEEEtran.bst: No hyphenation pattern has been}%
\typeout{** loaded for the language `#1'. Using the pattern for}%
\typeout{** the default language instead.}%
\else
\language=\csname l@#1\endcsname
\fi
#2}}
\providecommand{\BIBdecl}{\relax}
\BIBdecl

\bibitem{UCRArchive2018}
\BIBentryALTinterwordspacing
H.~A. Dau, E.~Keogh, K.~Kamgar, C.-C.~M. Yeh, Y.~Zhu, S.~Gharghabi, C.~A.
  Ratanamahatana, Yanping, B.~Hu, N.~Begum, A.~Bagnall, A.~Mueen, G.~Batista,
  and Hexagon-ML. (2018, October) The ucr time series classification archive.
  [Online]. Available:
  \url{https://www.cs.ucr.edu/~eamonn/time\_series\_data\_2018/}
\BIBentrySTDinterwordspacing

\bibitem{mueen_time_2014}
\BIBentryALTinterwordspacing
A.~Mueen, ``\BIBforeignlanguage{en}{Time series motif discovery: dimensions and
  applications},'' \emph{\BIBforeignlanguage{en}{Wiley Interdisciplinary
  Reviews: Data Mining and Knowledge Discovery}}, vol.~4, no.~2, pp. 152--159,
  2014. [Online]. Available:
  \url{https://onlinelibrary.wiley.com/doi/abs/10.1002/widm.1119}
\BIBentrySTDinterwordspacing

\bibitem{hao_visual_2012}
\BIBentryALTinterwordspacing
M.~C. Hao, M.~Marwah, H.~Janetzko, U.~Dayal, D.~A. Keim, D.~Patnaik,
  N.~Ramakrishnan, and R.~K. Sharma, ``\BIBforeignlanguage{en}{Visual
  exploration of frequent patterns in multivariate time series},''
  \emph{\BIBforeignlanguage{en}{Information Visualization}}, vol.~11, no.~1,
  pp. 71--83, Jan. 2012. [Online]. Available:
  \url{https://journals.sagepub.com/doi/abs/10.1177/1473871611430769}
\BIBentrySTDinterwordspacing

\bibitem{balasubramanian_flexible_2013}
A.~Balasubramanian and B.~Prabhakaran, ``Flexible exploration and visualization
  of motifs in biomedical sensor data,'' in \emph{Proc. of {Workshop} on {Data}
  {Mining} for {Healthcare}, in conjunction with {ACM} {KDD}}, 2013.

\bibitem{senin_grammarviz_2014}
P.~Senin, J.~Lin, X.~Wang, T.~Oates, S.~Gandhi, A.~P. Boedihardjo, C.~Chen,
  S.~Frankenstein, and M.~Lerner, ``\BIBforeignlanguage{en}{{GrammarViz} 2.0:
  {A} {Tool} for {Grammar}-{Based} {Pattern} {Discovery} in {Time} {Series}},''
  in \emph{\BIBforeignlanguage{en}{Machine {Learning} and {Knowledge}
  {Discovery} in {Databases}}}, ser. Lecture {Notes} in {Computer} {Science},
  T.~Calders, F.~Esposito, E.~Hüllermeier, and R.~Meo, Eds.\hskip 1em plus
  0.5em minus 0.4em\relax Berlin, Heidelberg: Springer, 2014, pp. 468--472.

\bibitem{Kohonen2001}
T.~Kohonen, \emph{{Self-Organizing Maps}}, 3rd~ed.\hskip 1em plus 0.5em minus
  0.4em\relax Berlin: Springer, 2001.

\bibitem{berndt_using_1994}
D.~J. Berndt and J.~Clifford, ``Using {Dynamic} {Time} {Warping} to {Find}
  {Patterns} in {Time} {Series},'' in \emph{Proceedings of the {AAAI}
  {Workshop} on {Knowledge} {Discovery} in {Databases}}, 1994, pp. 359--370.

\bibitem{tanaka_discovery_2005}
\BIBentryALTinterwordspacing
Y.~Tanaka, K.~Iwamoto, and K.~Uehara, ``\BIBforeignlanguage{en}{Discovery of
  {Time}-{Series} {Motif} from {Multi}-{Dimensional} {Data} {Based} on {MDL}
  {Principle}},'' \emph{\BIBforeignlanguage{en}{Machine Learning}}, vol.~58,
  no.~2, pp. 269--300, Feb. 2005. [Online]. Available:
  \url{https://doi.org/10.1007/s10994-005-5829-2}
\BIBentrySTDinterwordspacing

\bibitem{torkamani_survey_2017}
\BIBentryALTinterwordspacing
S.~Torkamani and V.~Lohweg, ``\BIBforeignlanguage{en}{Survey on time series
  motif discovery},'' \emph{\BIBforeignlanguage{en}{Wiley Interdisciplinary
  Reviews: Data Mining and Knowledge Discovery}}, vol.~7, no.~2, p. e1199,
  2017. [Online]. Available:
  \url{https://onlinelibrary.wiley.com/doi/abs/10.1002/widm.1199}
\BIBentrySTDinterwordspacing

\bibitem{lin_finding_2002}
\BIBentryALTinterwordspacing
J.~Lin, E.~Keogh, S.~Lonardi, and P.~Patel, ``Finding {Motifs} in {Time}
  {Series},'' in \emph{Proceedings of the {Second} {Workshop} on {Temporal}
  {Data} {Mining}}, Jul. 2002, pp. 53--68. [Online]. Available:
  \url{http://citeseer.ist.psu.edu/lin02finding.html}
\BIBentrySTDinterwordspacing

\bibitem{lin_experiencing_2007}
\BIBentryALTinterwordspacing
J.~Lin, E.~Keogh, L.~Wei, and S.~Lonardi,
  ``\BIBforeignlanguage{en}{Experiencing {SAX}: a novel symbolic representation
  of time series},'' \emph{\BIBforeignlanguage{en}{Data Mining and Knowledge
  Discovery}}, vol.~15, no.~2, pp. 107--144, Oct. 2007. [Online]. Available:
  \url{https://link.springer.com/article/10.1007/s10618-007-0064-z}
\BIBentrySTDinterwordspacing

\bibitem{das_rule_1998}
\BIBentryALTinterwordspacing
G.~Das, K.-I. Lin, H.~Mannila, G.~Renganathan, and P.~Smyth, ``Rule {Discovery}
  from {Time} {Series},'' in \emph{Proceedings of the {Fourth} {International}
  {Conference} on {Knowledge} {Discovery} and {Data} {Mining}}, ser.
  {KDD}'98.\hskip 1em plus 0.5em minus 0.4em\relax New York, NY: AAAI Press,
  1998, pp. 16--22. [Online]. Available:
  \url{http://dl.acm.org/citation.cfm?id=3000292.3000296}
\BIBentrySTDinterwordspacing

\bibitem{mueen_exact_2009}
\BIBentryALTinterwordspacing
A.~Mueen, E.~Keogh, Q.~Zhu, S.~Cash, and B.~Westover, ``Exact {Discovery} of
  {Time} {Series} {Motifs},'' in \emph{Proceedings of the 2009 {SIAM}
  {International} {Conference} on {Data} {Mining}}, ser. Proceedings.\hskip 1em
  plus 0.5em minus 0.4em\relax Society for Industrial and Applied Mathematics,
  Apr. 2009, pp. 473--484. [Online]. Available:
  \url{https://epubs.siam.org/doi/abs/10.1137/1.9781611972795.41}
\BIBentrySTDinterwordspacing

\bibitem{yeh_matrix_2016}
C.-C.~M. Yeh, Y.~Zhu, L.~Ulanova, N.~Begum, Y.~Ding, H.~A. Dau, D.~F. Silva,
  A.~Mueen, and E.~Keogh, ``Matrix {Profile} {I}: {All} {Pairs} {Similarity}
  {Joins} for {Time} {Series}: {A} {Unifying} {View} {That} {Includes}
  {Motifs}, {Discords} and {Shapelets},'' in \emph{2016 {IEEE} 16th
  {International} {Conference} on {Data} {Mining} ({ICDM})}, Dec. 2016, pp.
  1317--1322, iSSN: 2374-8486.

\bibitem{nunthanid_parameter-free_2012}
P.~Nunthanid, V.~Niennattrakul, and C.~A. Ratanamahatana, ``Parameter-free
  motif discovery for time series data,'' in \emph{2012 9th {International}
  {Conference} on {Electrical} {Engineering}/{Electronics}, {Computer},
  {Telecommunications} and {Information} {Technology}}, May 2012, pp. 1--4.

\bibitem{gao_exploring_2018}
\BIBentryALTinterwordspacing
Y.~Gao and J.~Lin, ``\BIBforeignlanguage{en}{Exploring variable-length time
  series motifs in one hundred million length scale},''
  \emph{\BIBforeignlanguage{en}{Data Mining and Knowledge Discovery}}, vol.~32,
  no.~5, pp. 1200--1228, Sep. 2018. [Online]. Available:
  \url{https://doi.org/10.1007/s10618-018-0570-1}
\BIBentrySTDinterwordspacing

\bibitem{lin_finding_2010}
J.~Lin and Y.~Li, ``Finding approximate frequent patterns in streaming medical
  data,'' in \emph{2010 {IEEE} 23rd {International} {Symposium} on
  {Computer}-{Based} {Medical} {Systems} ({CBMS})}, Oct. 2010, pp. 13--18,
  iSSN: 1063-7125.

\bibitem{Henriques2010}
\BIBentryALTinterwordspacing
R.~Henriques, F.~Bacao, and V.~Lobo, ``{Artificial Intelligence in Geospatial
  Analysis: applications of Self-Organizing Maps in the context of Geographic
  Information Science},'' Ph.D. dissertation, Lisboa, 2010. [Online].
  Available: \url{http://hdl.handle.net/10362/5723}
\BIBentrySTDinterwordspacing

\bibitem{Kaski1996}
\BIBentryALTinterwordspacing
S.~Kaski and T.~Kohonen, ``{Exploratory data analysis by the self-organizing
  map: structures of welfare and poverty in the world},'' in \emph{Neural
  Networks in Financial Engineering}, N.~Apostolos-Paul, {Yaser Refenes},
  {Yaser Abu-Mostafa}, {John Moody}, and A.~Weigend, Eds.\hskip 1em plus 0.5em
  minus 0.4em\relax Singapore: World Scientific, 1996, pp. 498--507. [Online].
  Available: \url{http://www.cis.hut.fi/{~}sami/therest.html}
\BIBentrySTDinterwordspacing

\bibitem{Kaski1998}
\BIBentryALTinterwordspacing
S.~Kaski, J.~Nikkil{\"{a}}, and T.~Kohonen, ``{Methods for interpreting a
  self-organized map in data analysis},'' in \emph{Proceedings of ESANN'98 ,
  6th European Symposium on Artificial Neural Networks}, M.~Verleysen,
  Ed.\hskip 1em plus 0.5em minus 0.4em\relax Bruges, Belgium: D-Facto, 1998,
  pp. 185--190. [Online]. Available:
  \url{http://www.cis.hut.fi/{~}sami/therest.html}
\BIBentrySTDinterwordspacing

\bibitem{Henriques2009}
\BIBentryALTinterwordspacing
R.~Henriques, F.~Ba{\c{C}}{\~{a}}o, and V.~Lobo, ``{Carto-SOM: cartogram
  creation using self-organizing maps},'' \emph{International Journal of
  Geographical Information Science}, vol.~23, no.~4, pp. 483--511, 2009.
  [Online]. Available:
  \url{https://www.tandfonline.com/doi/abs/10.1080/13658810801958885}
\BIBentrySTDinterwordspacing

\bibitem{Vesanto1999}
\BIBentryALTinterwordspacing
J.~Vesanto, ``{SOM-based data visualization methods},'' \emph{Intelligent Data
  Analysis}, vol.~3, pp. 111--126, 1999. [Online]. Available:
  \url{https://dl.acm.org/doi/10.1016/S1088-467X\%2899\%2900013-X}
\BIBentrySTDinterwordspacing

\bibitem{Ultsch1990}
A.~Ultsch and H.~P. Siemon, ``{Kohonen Networks on Transputers: Implementation
  and Animation},'' in \emph{International Neural Network Conference (INNC)},
  1990.

\bibitem{wang2013}
D.~{Wang} and S.~{Tapan}, ``A robust elicitation algorithm for discovering dna
  motifs using fuzzy self-organizing maps,'' \emph{IEEE Transactions on Neural
  Networks and Learning Systems}, vol.~24, no.~10, pp. 1677--1688, Oct 2013.

\bibitem{bassani2015dimension}
H.~F. Bassani and A.~F.~R. Araujo, ``{Dimension Selective Self-Organizing Maps
  With Time-Varying Structure for Subspace and Projected Clustering},''
  \emph{IEEE Transactions on Neural Networks and Learning Systems}, vol.~26,
  no.~3, pp. 458--471, mar 2015.

\bibitem{Brito_2018}
\BIBentryALTinterwordspacing
R.~C. Brito and H.~F. Bassani, ``Self-organizing maps with variable input
  length for motif discovery and word segmentation,'' \emph{2018 International
  Joint Conference on Neural Networks (IJCNN)}, Jul 2018. [Online]. Available:
  \url{http://dx.doi.org/10.1109/IJCNN.2018.8489090}
\BIBentrySTDinterwordspacing

\bibitem{Novikov2019}
\BIBentryALTinterwordspacing
A.~Novikov, ``{PyClustering}: Data mining library,'' \emph{Journal of Open
  Source Software}, vol.~4, no.~36, p. 1230, apr 2019. [Online]. Available:
  \url{https://doi.org/10.21105/joss.01230}
\BIBentrySTDinterwordspacing

\bibitem{Ratanamahatana2004}
\BIBentryALTinterwordspacing
C.~A. Ratanamahatana and E.~Keogh, ``Making time-series classification more
  accurate using learned constraints,'' in \emph{Proceedings of the 2004 {SIAM}
  International Conference on Data Mining}.\hskip 1em plus 0.5em minus
  0.4em\relax Society for Industrial and Applied Mathematics, Apr. 2004.
  [Online]. Available: \url{https://doi.org/10.1137/1.9781611972740.2}
\BIBentrySTDinterwordspacing

\bibitem{liu_uwave_2009}
\BIBentryALTinterwordspacing
J.~Liu, L.~Zhong, J.~Wickramasuriya, and V.~Vasudevan,
  ``\BIBforeignlanguage{en}{{uWave}: {Accelerometer}-based personalized gesture
  recognition and its applications},'' \emph{\BIBforeignlanguage{en}{Pervasive
  and Mobile Computing}}, vol.~5, no.~6, pp. 657--675, Dec. 2009. [Online].
  Available:
  \url{http://www.sciencedirect.com/science/article/pii/S1574119209000674}
\BIBentrySTDinterwordspacing

\end{thebibliography}

\end{document}